\documentclass{article}
\usepackage{spconf,amsmath,graphicx}

\title{Multiview Variational Graph Autoencoders\\ for Canonical Correlation Analysis}
\name{Yacouba Kaloga$^\S$, Pierre Borgnat$^\S$, Sundeep Prabhakar Chepuri$^\star$, Patrice Abry$^\S$ and Amaury Habrard$^\dag$\thanks{ Supported by the IFCAM project MA/IFCAM/19/56, SERB SRG/2019/000619 grant, the ACADEMICS Grant of IDEXLYON, Univ. Lyon, PIA  ANR-16-IDEX-0005, and the CBP IT test platform (ENS de Lyon, France) for ML facilities,
operating the SIDUS solution~\cite{SIDUS}.}}
\address{$^\S$Univ Lyon, Ens de Lyon, Univ. Claude Bernard, CNRS, Laboratoire de Physique, Lyon, France\\ 
$^\star$Department of Electrical and Communication Engineering, Indian Institute of Science, Bangalore, India \\
$^\dag$ University of Lyon, UJM-Saint-Etienne, CNRS, Laboratoire Hubert Curien, UMR 5516, France
}
\usepackage{xcolor}

\usepackage{amssymb}
\usepackage{amsmath}
\usepackage{url}
\usepackage{booktabs}

\usepackage{pifont}
\usepackage{array}
\newcolumntype{h}{>{\setbox0=\hbox\bgroup}c<{\egroup}@{}}

\def\cast{{
   \mathord{
      \hbox to 0em{
         \ooalign{
	   \smash{\hbox{$\ast$}}\crcr
	   \smash{\hskip-1pt\Large\hbox{$\circ$}} }
	 \hidewidth}
      \phantom{\bigcirc}
} }}

\newcommand{\bds}{\begin {itemize}}
\newcommand{\eds}{\end {itemize}}
\newcommand{\bdf}{\begin{definition}}
\newcommand{\blm}{\begin{lemma}}
\newcommand{\edf}{\end{definition}}
\newcommand{\elm}{\end{lemma}}
\newcommand{\bthm}{\begin{theorem}}
\newcommand{\ethm}{\end{theorem}}
\newcommand{\bprp}{\begin{prop}}
\newcommand{\eprp}{\end{prop}}
\newcommand{\bcl}{\begin{claim}}
\newcommand{\ecl}{\end{claim}}
\newcommand{\bcr}{\begin{coro}}
\newcommand{\ecr}{\end{coro}}
\newcommand{\bquest}{\begin{question}}
\newcommand{\equest}{\end{question}}

\newcommand{\larrow}{{\larrow}}

\newcommand{\argmin}{\ensuremath{\mathrm{arg}\min}}
\newcommand{\argmax}{\ensuremath{\mathrm{arg}\max}}

\usepackage{latexsym}

\def\IC{\mathbb C}
\def\IN{\mathbb N}
\def\IZ{\mathbb Z}
\def\IR{\mathbb R}

\def\shat{^{\mathchoice{}{}%
 {\,\,\smash{\hbox{\lower4pt\hbox{$\widehat{\null}$}}}}%
 {\,\smash{\hbox{\lower3pt\hbox{$\hat{\null}$}}}}}}

\def\bSigma{{
      \ooalign{
      \smash{\hskip.4pt\raise.4pt\hbox{$\Sigma$}}\vphantom{}\crcr
      \smash{\hskip.7pt\raise.6pt\hbox{$\Sigma$}}\vphantom{}\crcr
      \smash{\hbox{$\Sigma$}}\vphantom{$\Sigma$}}
      \vphantom{\hbox{$\Sigma$}}
      }}
\def\bTheta{{
      \ooalign{
      \smash{\hskip.5pt\raise.5pt\hbox{$\Theta$}}\vphantom{}\crcr
      \smash{\hskip.0pt\raise.1pt\hbox{$\Theta$}}\vphantom{}\crcr
      \smash{\hbox{$\Theta$}}\vphantom{$\Theta$}}
      \vphantom{\hbox{$\Theta$}}
      }}
\def\bDelta{{
      \ooalign{
      \smash{\hskip.4pt\raise.4pt\hbox{$\Delta$}}\vphantom{}\crcr
      \smash{\hskip.7pt\raise.6pt\hbox{$\Delta$}}\vphantom{}\crcr
      \smash{\hbox{$\Delta$}}\vphantom{$\Delta$}}
      \vphantom{\hbox{$\Delta$}}
      }}
\def\bLambda{{
      \ooalign{
      \smash{\hskip.5pt\raise.5pt\hbox{$\Lambda$}}\vphantom{}\crcr
      \smash{\hskip.0pt\raise.1pt\hbox{$\Lambda$}}\vphantom{}\crcr
      \smash{\hbox{$\Lambda$}}\vphantom{$\Lambda$}}
      \vphantom{\hbox{$\Lambda$}}
      }}

\makeatletter

\def\bordermatrix#1{\begingroup \m@th
  \@tempdima 8.75\p@
  \setbox\z@\vbox{%
    \def\cr{\crcr\noalign{\kern2\p@\global\let\cr\endline}}%
    \ialign{$##$\hfil\kern2\p@\kern\@tempdima&\thinspace\hfil$##$\hfil
      &&\quad\hfil$##$\hfil\crcr
      \omit\strut\hfil\crcr\noalign{\kern-\baselineskip}%
      #1\crcr\omit\strut\cr}}%
  \setbox\tw@\vbox{\unvcopy\z@\global\setbox\@ne\lastbox}%
  \setbox\tw@\hbox{\unhbox\@ne\unskip\global\setbox\@ne\lastbox}%
  \setbox\tw@\hbox{$\kern\wd\@ne\kern-\@tempdima\left[\kern-\wd\@ne
    \global\setbox\@ne\vbox{\box\@ne\kern2\p@}%
    \vcenter{\kern-\ht\@ne\unvbox\z@\kern-\baselineskip}\,\right]$}%
  \null\;\vbox{\kern\ht\@ne\box\tw@}\endgroup}
\makeatother

\makeatletter
\def\argmin{\mathop{\operator@font arg\,min}}
\def\argmax{\mathop{\operator@font arg\,max}}
\makeatother

\newcommand{\bea}{\begin{array}}
\newcommand{\ena}{\end{array}}
\newcommand{\beq}{\begin{equation}}
\newcommand{\enq}{\end{equation}}

\newcommand{\beqa}{\begin{eqnarray}}
\newcommand{\enqa}{\end{eqnarray}}

\newcommand{\beqan}{\begin{eqnarray*}}
\newcommand{\enqan}{\end{eqnarray*}}

\newcommand{\AL}{\begin{enumerate}}
\newcommand{\ALE}{\end{enumerate}}

\def\addots{\mathinner{
    \mkern1mu\raise0pt\vbox{\kern7pt\hbox{.}}
    \mkern2mu\raise4pt\hbox{.}
    \mkern2mu\raise7pt\hbox{.}
    \mkern1mu}}

\def\sddots{\mathinner{
    \mkern.8mu\raise7pt\hbox{.}
    \mkern.8mu\raise4pt\hbox{.}
    \mkern.8mu\raise0pt\vbox{\kern7pt\hbox{.}}
    \mkern1mu}}

\def\saddots{\mathinner{
    \mkern.2mu\raise0pt\vbox{\kern7pt\hbox{.}}
    \mkern.2mu\raise4pt\hbox{.}
    \mkern.2mu\raise7pt\hbox{.}
    \mkern1mu}}

\def\sqplus{\mathbin{
	{\ooalign{\hfil\raise.3ex\hbox{\scriptsize
	+}\hfil\crcr\mathhexbox274\crcr\mathhexbox275}}
	}} 
\def\sqminus{\mathbin{
	{\ooalign{\hfil\raise.3ex\hbox{\scriptsize
	--}\hfil\crcr\mathhexbox274\crcr\mathhexbox275}}
	}}

\def\IC{{
   \mathord{
      \hbox to 0em{
	 \hskip-4pt
         \ooalign{
	   \smash{\hskip1.9pt\raise2.6pt\hbox{$\scriptscriptstyle |$}}\crcr
	   \smash{\hbox{\rm\sf C}} }
	 \hidewidth}
      \phantom{\hbox{\rm\sf C}}
} }}
\def\IN{
    {\ooalign{
   \smash{\hskip2.2pt\raise1.5pt\hbox{$\scriptscriptstyle |$}}\vphantom{}\crcr
   \hbox{\sf N}
	}}
	} 
\def\IZ{
    {\ooalign{
   \smash{\hskip1.9pt\raise0pt\hbox{$\sf Z$}}\vphantom{}\crcr
   \hbox{\sf Z}
	}}
	} 
\def\IR{
    {\ooalign{
   \smash{\hskip2.2pt\raise1.5pt\hbox{$\scriptscriptstyle |$}}\vphantom{}\crcr
   \smash{\hskip2.2pt\raise3.3pt\hbox{$\scriptscriptstyle |$}}\vphantom{}\crcr
   \hbox{\sf R}
	}}
	} 

\DeclareMathAlphabet{\mathcmb}{OT1}{cmr}{b}{n}

\def\bSigma{\ensuremath{\mathcmb{\Sigma}}}
\def\bLambda{\ensuremath{\mathcmb{\Lambda}}}

\def\bTheta{\ensuremath{\mathcmb{\Theta}}}

\newcommand{\SI}{\begin{indlist}}
\newcommand{\EI}{\end{indlist}}

\newcommand{\DL}{\begin{dashlist}}
\newcommand{\DLE}{\end{dashlist}}

\makeatletter
\def\setboxz@h{\setbox\z@\hbox}
\def\wdz@{\wd\z@}
\def\boxz@{\box\z@}
\def\underset#1#2{\binrel@{#2}%
  \binrel@@{\mathop{\kern\z@#2}\limits_{#1}}}
\def\binrel@#1{\begingroup
  \setboxz@h{\thinmuskip0mu
    \medmuskip\m@ne mu\thickmuskip\@ne mu
    \setbox\tw@\hbox{$#1\m@th$}\kern-\wd\tw@
    ${}#1{}\m@th$}%
  \edef\@tempa{\endgroup\let\noexpand\binrel@@
    \ifdim\wdz@<\z@ \mathbin
    \else\ifdim\wdz@>\z@ \mathrel
    \else \relax\fi\fi}%
  \@tempa
}
\let\binrel@@\relax%
\makeatother

\begin{document}
%
\maketitle
\begin{abstract}
We present a novel multiview canonical correlation analysis model based on a variational approach. This is the first nonlinear model that takes into account the available graph-based geometric constraints  while being scalable for processing large scale  datasets with multiple views.  It is based on an autoencoder architecture with graph convolutional neural network layers. 
We experiment with our approach on classification, clustering, and recommendation tasks on real datasets. The algorithm is competitive with state-of-the-art multiview representation learning techniques.

\end{abstract}
\begin{keywords}
Canonical correlation analysis, dimensionality reduction,  multiview, graph neurals networks, variational inference
\end{keywords}
\section{Introduction}
\label{sec:intro}

Interconnected societies generate large amounts of structured data, that frequently stem from observing a common set of objects (or sources) through different modalities. Such multiview datasets are encountered in many different fields like computational biology \cite{Yam03}, acoustics~\cite{Arora13}, surveillance~\cite{Col01}, or social networks~\cite{Benton:2016}, to list a few. Although there exist many tools to analyze and study multiview datasets~\cite{Li19},  it is more than ever necessary to develop methods for analyzing large-scale and structured multiview datasets.

Canonical Correlation Analysis (CCA) \cite{Hotel36} can be used for multiview representation learning, by seeking latent low-dimensional representations that are common to all the different views. 
This common representation that encodes information from different datasets can be leveraged to improve the performance of machine learning tasks, e.g., clustering \cite{Chaud09}.
\emph{Algebraic} approaches to CCA obtain this latent low-dimensional manifold by maximizing correlations between the projections on the different views onto it. Being nonparametric, these approaches are powerful but do not scale well to large datasets. 
Instead, \emph{probabilistic} approaches to CCA scale more easily 
but, being based on models, are less versatile to adjust to model mismatch or to the data structure. The present work attempts to reconcile scalability and versatility in CCA.

It has then been extended to the multiview setting, e.g., see \cite{Kettenring71}, and to account for nonlinear dependencies (beyond  correlations), 
e.g., see Kernel CCA \cite{Aka01}, Deep CCA \cite{Andre13}, or Autoencoder CCA \cite{Wang15}. Despite significant improvements in performance, 
many of these approaches suffer from scalability issues~\cite{Chang18,lopez14}, mainly due to the prohibitive costs of the underlying eigendecomposition. 
Alternatively, it has been shown that CCA can be cast equivalently into a Bayesian inference problem \cite{Bach05}.
As recent advances in Variational autoencoders \cite{Kipf16} made Bayesian inference scalable, probabilistic CCA approaches gained popularity because of their potential 
(e.g., to deal with missing views) and scalability. 
Concomitantly,  it was shown in~\cite{Chen18,Chen19} that incorporating the available graph-induced knowledge about the common source into multiview CCA improves performance of various machine learning tasks. We refer to this graph-aware multiview CCA method from~\cite{Chen19} as GMCCA. However, GMCCA suffers from the involved eigendecomposition costs. 
However, there is no CCA method that both can use some prior graph-based structure in the latent space, and is scalable. Such a method is proposed here. 

The present work develops a scalable multiview variational graph autoencoder for CCA (MVGCCA). 
Section~\ref{sec:background} recalls preliminaries in Multiview CCA. 
Section~\ref{sec:proposition} describes the proposed approach and its key contributions. 
In particular, the way in which the graph structure is enforced in the common latent space preserving the scalability.
Section~\ref{sec:data} describes the datasets that are used for numerical experiments in Section~\ref{sec:results}.


\section{Multiview-CCA Background}
\label{sec:background}

\noindent{\bf Setting.} We consider $M$-view datasets where the instance space $X$ has $M$ view, spaces in $\mathbb{R}^{d_m}$, $m=1,\ldots,M$. We denote $n$ the sample size, and $X_m \in \mathbb{R}^{d_m \times n}$ the $m$-view data matrix. An instance $i$ of view $m$ is denoted as $x_m ^i \in  \mathbb{R}^{d_m}$.


\subsection{Algebraic approaches: linear CCA and    GMCCA}

Given 2-views, seek the best projectors $U_1 \in \mathbb{R}^{d \times d_1}$, $U_2\in \mathbb{R}^{d \times d_2}$ with $d <\!\!< \min{(d_1, d_2)}$ such that the correlation between $U_1 X_1$ and $U_2 X_2$ is maximized.
This can be formulated as the following optimization problem:
\begin{equation}
\label{cca_objective}
    \begin{split}
    \min_{U_1,U_2}  |\!|U_1 X_1 - U_2 X_2  |\!|_2^2
     &\text{\:\:\:s.t.\:\:\:} U_m(X_m X_m^{T})U_m^{T} = I_{d_m}.
    \end{split} 
\end{equation} 
The solution is obtained via an eigendecomposition (e.g., see \cite{Andre13,Chen19}).
Extending this to multiview data is difficult as maximizing the pairwise correlation between the $M$ views is NP-hard \cite{Rup13}. A relaxation is to seek a common low-dimensional representation $S \in \mathbb{R}^{d \times n}$, that is as close as possible to each low-dimensional projection $U_mX_m, m = 1 \dots M$. It leads to problem~(\ref{mcca_objective}) with $\gamma =0$. This relaxation is solved using an eigenvalue decomposition. Chen et al. \cite{Chen19} have proposed GMCCA, in which some graph-based prior knowledge on $S$, when available, can be incorporated to increase the clustering performance. This is done by ensuring smoothness of $S$ on this graph. By doing so, graph-regularized CCA problem can be posed as (with $L$ the Laplacian matrix):
%
%
\begin{equation}
\label{mcca_objective}
    \min_{(U_m)_{m=1 \dots M}}  \sum_{ \substack{m=1}  }^M |\! |U_m X_m - S  |\!|_2^2+ \gamma SLS^{T}
     \text{\:\:\:s.t.\:\:\:} SS^{T} =  {I}_{d}.
\end{equation} 
The solution $S^{*}$ has columns equal to the $k$ leading eigenvectors of $\sum_{m=1}^{M} X_m^{T}(X_m X_m^{T})X_m - \gamma L$ \cite{Chen18}.
For large datasets, this method involving the eigendecomposition does not always scale well.

\subsection{Probabilistic CCA}
The CCA solution of eq.~(\ref{cca_objective}) can be obtained from a graphical model~\cite{Bach05}, where the views come from a common source $Z$. Let us define a prior distribution on this latent space $p(z)$, and the conditional probability (decoders) for each views as $p_{\theta_m}(x_1|z)$, and are given as\footnote{Notation $p(\cdot) \sim q$, means $p = q$ .}:
\begin{equation}
    \label{ccabach}
    \begin{split}
         p(z) &\sim \mathcal{N}(0,I_d). \\
         p_{\theta_m}(x_m|z) &\sim \mathcal{N}(W_m \cdot z + \mu_m,\Psi_m ).
    \end{split}
\end{equation}
with $\mu_m \in \mathbb{R}^{d_m}$, $W_m \in \mathbb{R}^{d_m \times d}$ and $\Psi_m \in \mathbb{R}^{d_m \times d_m} \succcurlyeq 0$.  We collect  the parameters in $\theta_m$ as $\theta_m =(W_m, \mu_m, \Psi_m)$, $m=1,2$. The optimal solution to the probabilistic two-views CCA $(\theta_1^*,\theta_2^*)$ is computed by maximizing data log-likelihood\footnote{$\log p_\theta(X_1,X_2)  = \sum_{i = 1}^n{ \sum_{m=1}^2 \log \int_{\mathbb{R}^d }p_{\theta_m}( x_m^i | z)  p(z) dz}$}. From the Bayes theorem, the probabilities (defining encoders), $P_{\theta_1^{*}}(z|x^i_1)$ and $P_{\theta_2^{*}}(z|x^i_2)$ are well defined.
Their expectation is exactly the optimal projection (up to an operator $M_m\in \mathbb{R}^{ d \times d}$ defined in \cite{Bach05}): $\mathbb{E}_{z \sim p}P(z|x^i_m) = M_m^TU_m^*x^i_m$, $m=1,2$. 
In this framework, CCA has a natural multiview extension to $M>2$. We will use such an extension, while incorporating graph regularization like in \cite{Chen19}.
%
%
%
%
Yet, solving a problem such as eq.~(\ref{ccabach}) (i.e., a multi-dimensional probability distribution) is often intractable because it requires maximization of the log-likelihood and thus to integrate over all the latent spaces. A variational approach solves this issue.

\subsection{Variational bound and graph autoencoder}
Kingma et al.~\cite{King14} have shown that introducing parametric distributions $q_{\eta}(z|x)$, with parameters $\eta$, of some untractable distribution $p(z|x)$, one can build a lower bound of log-likelihood called evidence lower bound objective (ELBO)\footnote{Notation $z \sim p$, means $z$ follows distribution $p$.}: 
%
%
%
\begin{equation}
    \label{elbo1}
    \begin{split}
        \log p_\theta(X_1, X_2) \geq  & \sum_{i = 1}^{n} \mathbb{E}_{z \sim q_\eta(\cdot|x_1^i, x_2^i)}[\log(p_\theta(x_1^i,x_2^i|z)) ] \\ & - \sum_{i = 1}^{n}  D_{KL}(q_\eta(\cdot|x_1^i, x_2^i)|\!|p). 
    \end{split}
\end{equation}
ELBO is easier to approximate than the data log-likelihood so we maximise this lower bound with respect to both $\theta$ and $\eta$. The first term of ELBO ensures a correct data reconstruction due to encoded latent representations. The second term acts as a regularizer ensuring that posterior distribution $q_\eta$ for each multiview element remains coherent in the latent space. It corresponds to the loss function of the variational autoencoders, used in most existing variational CCA methods.

It is possible to account geometric structure, by using the variational autoencoder extension proposed by Kipf et al. \cite{Kipf16} for link prediction on graphs. In their singleview framework, data resides on the nodes of a graph with the weight matrix $A \in [0,1]^{n \times n}$. We denote $\mathcal{V}(x_1^i)$ the neighborhood  of $x_1^i$  (including $x_1^i$)  up to a certain distance in the graph. Then the graph-aware ELBO loss function is defined as :
\begin{equation}
    \label{elbokipf}
    \begin{split}
        \mathcal{L}_{ELBO} &= \sum_{ i,j = 1}^n  \mathbb{E}_{ \substack{ z^i \sim q_\eta(\cdot|\mathcal{V}(x_1^i),A) \\ z^j \sim q_\eta(\cdot|\mathcal{V}(x_1^j),A) }}
        [\log(p_\theta(A_{i,j}|z^i,z^j))]  \\ & -  \sum_{i = 1}^n  D_{KL}(q_\eta(\cdot| \mathcal{V}(x_1^i),A)|\!|p).
    \end{split}
\end{equation}
The cross entropy term 
allows for graph reconstruction~\cite{Kipf16}. The posterior probability $q_\eta$ can be now parametrized as a graph neural network.
We will use a similar approach here, and extend it for multiple views.

\section{Variational Graph M-CCA}
\label{sec:proposition}

\begin{figure}[t!]
    \centering
\includegraphics[width=5cm,height=29cm,keepaspectratio]{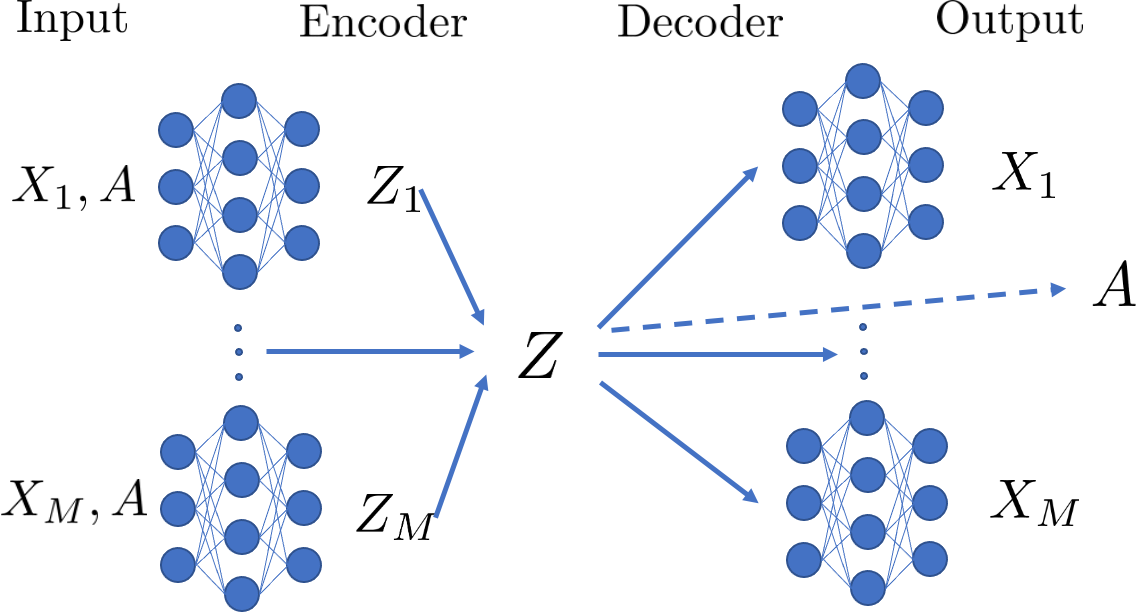}
    \caption{Representation of MVGCCA. 
    All the views are encoded to their own latent space $Z_m$ using the common graph. They are merged to form a common view $Z$. Finally, $Z$ is tailored to  decode all the views and original graph.}
    \label{fig:test}
    \vspace{-2mm}
\end{figure}

We now present the proposed method that estimate a probabilistic Multiview-CCA. Starting from eq.~(\ref{ccabach}), our first contribution is to simply extend this framework for $m > 2$. To do so, we model the $M$-view data given the latent space vector $z \in \mathbb{R}^{d} \sim p  = \mathcal{N}(0,I_d)$ using a parametric decoder based on a multilayer perceptron (denoted \textbf{MLP}). Hence, for $m=1,2,\ldots,M$, we have $p_{\theta_m}(x_m|z)  \sim \mathcal{N}( {\boldsymbol{W}_m ^\mu}^{\text{dec}}\textbf{MLP}_m(z),\boldsymbol{\Psi_m} )$.
%

Next, we take into account a prior graph structure on the latent space of $z$, inspired by~\cite{Kipf16}: a link decoder distribution $p_g(a |z,z')\sim \mathcal{B}(a,\ell(z^T z'))$ is introduced, parametrized by a continuous Bernoulli law $\mathcal{B}$\footnote{$\mathcal{B}(a,p) = p^a(1-p)^{1-a}$. },  and $\ell(\cdot)$ is the logistic sigmoid function. Here $a\in[0,1]$ is a weight space vector and $z' \in \mathbb{R}^{d}$ is second latent space vector.

With the hypothesis of independence of all the views and of the links in the graph, we have the factorized decoder distribution  $p_\theta((x_m^i)_{m=1}^M|z) = \prod_{\substack{m =1}}^{M}p_\theta(x_m^i|z)$, graph decoder distribution\footnote{with $Z \in \mathbb{R}^{d \times n}$ with $Z(:,i) \sim q_{\eta}(\cdot|X,A)$  } $p_\theta(A|Z) = \prod_{\substack{i=1\\j=1}}^{n}p_g(A_{i,j}|z^i,z^j)$, and the parametric decoder distribution $q_\eta(z|( \mathcal{V}(x_m^i))_{m=1}^M,A) = \prod_{m = 1}^{M}q_{\eta_m}(z| \mathcal{V}(x_m^i),A)$. The ELBO takes the form\footnote{ $\log$ in $\log \sigma_m$ is applied element-wise on matrix $\sigma_m$.}:
\begin{equation}
    \label{elboour}
    \begin{split}
        \mathcal{L}_{ELBO} = \sum_{i,j = 1}^n  \mathbb{E}_{ \substack{ z^i \sim q_{\eta}(\cdot|(\mathcal{V}(x_m^i))_{m=1}^M,A) \\ z^j \sim q_\eta(\cdot|(\mathcal{V}(x_m^j))_{m=1}^M,A) }}
        \log p_g(A_{i,j}|z^i,z^j)  
        \\   + \sum_{i = 1}^{n} \sum_{m = 1}^{M} \mathbb{E}_{z  \sim q_{\eta}(\cdot|(\mathcal{V}(x_m^i))_{m=1}^M,A)}\log p_\theta(x_m^i|z)
        \\  - \sum_{i = 1}^n  D_{KL}( q_{\eta}(\cdot| (\mathcal{V}(x_m^i))_{m=1}^M,A)|\!|p).
    \end{split}
\end{equation}
The last term involves views encoders:
\begin{equation}
    \label{decoder}
    \begin{split}
    q_{{\eta}_m}( \cdot |\mathcal{V}(x_m^i),A) & = \mathcal{N}(\mu_m^{\text{enc}}(:,i),\text{diag}(\sigma_m^{\text{enc}}(:,i)).\\
        \mu_m^{\text{enc}} & = {\boldsymbol{W}_m^\mu}^{\text{enc}} \textbf{Krylov}_m(X_m,A). \\
        \log \sigma_m^{\text{enc}} & = {\boldsymbol{W}_m^\sigma}^{\text{enc}} \textbf{Krylov}_m(X_m,A).
    \end{split}
\end{equation}
We choose in this model to parametrize $q_\eta$ by the posterior distribution of a \textbf{Krylov} Graph convolutional neural networks (GCN) \cite{luan2019break}. The reason is that GCN are efficient to extract features information of a node considering its neighborhood~\cite{defferrard2016convolutional,Kipf17}. They have been widely used in node classification, node clustering, and other graph related tasks. Currently, many methods have similar performance \cite{wu19}, and since the choice is not critical, we simply use a truncated Krylov GCN achitecture \cite{luan2019break} which has been proven to have good properties when stacking across graph layers.

Finally, the trainable parameters of the models are the weights $\boldsymbol{\Psi_m}$,  ${\boldsymbol{W}_m^\mu}^{\text{dec}}$, ${\boldsymbol{W}_m^\sigma}^{\text{enc}}$, ${\boldsymbol{W}_m^\mu}^{\text{enc}}$, the parameters of the multilayer perceptron $\textbf{MLP}_m$ and of the Krylov GCN layers $\textbf{Krylov}_m$.
The training can be done with the ELBO as loss function, using a suitable optimization method. The latent representations for each view $x_m^i$ is $ \mathbb{E}_{ z \sim q_{{\eta}_m}( \cdot |\mathcal{V}(x_m^i),A)} =  {\mu_m ^{\text{enc}}}(:,i)$. And the common latent representation is given by
$$ \mathbb{E}_{ z \sim q_{\eta
}( \cdot |\mathcal{V}(x_m^i)_{m=1}^M,A)} = \bigg[\sum_{m=1}^{M} \frac{\mu_m ^{\text{enc}}(k,i)}{{\sigma_m^{\text{enc}}}^2(k,i)}/\sum_{m=1}^{M} \frac{1}{{\sigma_m^{\text{enc}}}^2(k,i)} \bigg]_{k=1}^d.$$ 
This common representation is used for the experiments.



\begin{figure}[t!]
    \centering
\includegraphics[width=4cm,height=29cm,keepaspectratio]{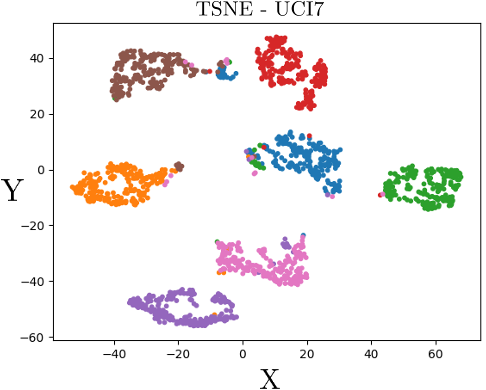}
    \caption{t-SNE visualisation in 2D of the latent space ($d=3)$ for the uci7 dataset. Each color represents a different class. 
      \label{fig:test}}
      \vspace{-2mm}
\end{figure}

\begin{table*}[t]
\centering
\scalebox{0.9}{
\begin{tabular}{|c||c|c|c||c|c|c||c|c|c||}
\hline
  Dataset     & \multicolumn{3}{c||}{\textbf{uci7}} & \multicolumn{3}{c||}{\textbf{uci10}} &  \multicolumn{3}{c||}{\begin{tabular}{@{}c@{}} \textbf{Recommendation}\end{tabular}} \\ \hline \hline 
Metric & Acc.     & ARI     & ARI2    & Acc.     & ARI     & ARI2   &      Prec.      & Recall       &      MRR  \\ \hline  \hline \hline 
PCA     &   0.87      &   0.55      & -    &   0.74       &   0.41      & -   &     0.1511     & 0.0795       &     0.3450   \\ \hline
GPCA    &    0.95    &    0.74    &  0.79      &         0.90 &     0.63    &    0.62  &     0.1578     & 0.0831       &    0.3649 \\ \hline
MCCA    &   0.89      &   0.66       &  -    &    0.79     &   0.59      &   -  &     0.0815     & 0.0429            &  0.2225  \\ \hline
GMCCA   &    0.96    &  \textbf{0.84}      &   0.83    &   0.92      &     \textbf{0.73}    &   \textbf{0.76}   &     \textbf{0.2290}     & \textbf{0.1206}             &  \textbf{ 0.4471 }  \\\hline 
MVGCCA   & \textbf{0.97}        &     0.79   &   \textbf{0.85}     &   \textbf{0.95}      &    0.69  &  \textbf{0.76} &    0.1951            &      0.0665       &    0.4230       \\ \hline
\end{tabular}
}

\caption{Results of experiments on the different datasets and tasks; see text for comments. MRR is mean reciprocal rank.
\label{data_results}}
\vspace{-2mm}
\end{table*}

\section{Datasets}
\label{sec:data}

\subsection{UCI handwritten datasets}
UCI Handwritten digits dataset\footnote{\scriptsize\url{archive.ics.uci.edu/ml/datasets/Multiple+Features}} is a multiview dataset of $n=2000$ small images representing digits. Each image has a label from 0 to 9 ($200$ elements for each) and 6 views of varying dimensions: $d_1 =76$, $d_2 =216$, $d_3 =64$, $d_4 =240$, $d_5 =47$ and $d_6 =6$; all views corresponding to specific transformations of the original image.
Clustering and classification tasks are performed on this dataset (uci10)
and on a partial version (uci7) where classes 0, 5 and 6 have been removed.

\subsection{Twitter friend recommendation}

A multiview dataset\footnote{\scriptsize\url{http://www.cs.jhu.edu/~mdredze/data/}} based on tweets from Twitter has been proposed in \cite{Benton:2016}. It consists of multiview representations of messages of users. There are $n = 102327$ users.
Each user has 6 1000-dimensional views: EgoTweets, MentionTweets, FriendTweets, FollowersTweets, FriendNetwork, and FollowerNetwork. 
A task of friend recommendation is performed as follows. The followed accounts are known for each user; given a highly followed account and a part of their followers, the goal is to determine, for each other users, whether or not he will follow this account after March 2015. For this task, a graph based on the Twitter dataset is built as in \cite{Chen19} with the views Egoweets, FollowersTweets, and FriendNetwork.
\label{sec:majhead}



\section{Experiments}
\label{sec:results}
All architectures and hyperparameters specified here have been fixed for all experiments unless otherwise indicated\footnote{Code available : \url{https://github.com/Yacnnn/MVGCCA}}. 
We preprocess all the views: each view is centered and normalized by its maximal absolute value. For each dataset graph adjacency matrix is rescaled with his maximal coefficient and diagonal coefficients are set to 1. \\
\textbf{Decoders:} the $\textbf{MLP}_m$ used as decoders have 4 hidden layers and 1024 units in each. 
We parametrize $\boldsymbol{\Psi}_m $ by a single scalar: $\boldsymbol{\Psi}_m = (\sigma_m^2 + 10^{-6})I$. This choice reduces complexity and improves robustness without decreasing performance.
\\
\textbf{Encoders:} the \textbf{Krylov} GCN layers \cite{luan2019break} encoding the mean and variance in $q_\eta$, looks up to 3 hop neighborhood; we stack 4 layers 
and use 1024 units in each hidden layer. \\
\textbf{General:}  Batch size is set to 512. A dropout regularization of rate 0.5 is also applied after all the hidden layers.  The Adam optimizer is always used for training.

\label{sec:majhead}

\subsection{UCI classification and clustering}

The model was trained on the uci7 and uci10 datasets. The loss function was the ELBO from (\ref{elboour}), trained in batches. For each batch, the graph used 
is the subgraph for the samples in the batch. The latent space is of dimension $d = 3$.  

Once having the latent representation of the elements in the uci dataset, 10-fold K-means clustering and spectral clustering are done. The quality of the clustering is evaluated with adjusted Rand index (ARI). A 10-fold SVM classification task is also performed. In order to select the best learning rate, number of epochs, dropout rate and K-means/SVM parameters, we performed 3 runs on different parameter combinations. The results are averaged on the 10-fold SVM classification. Finally we use the best performing parameters. 

Following \cite{Chen19}, the results are compared to PCA (applied on concatenated views), graph-regularized PCA, regular MCCA and GMCCA. In order to make fair comparison in terms of hyparameter tuning, all the experiments were done with the same protocol. The results are in Table~\ref{data_results}.

We see that MVGCCA is competitive in both classification (Acc.) and clustering tasks. It achieves the best performance on classification.  
For clustering, the quality with K-means (ARI1) on both datasets is slightly lower than the top state-of-the-art, while spectral clustering (ARI2) obtains the best score. This indicates that the graph structure seems to be well integrated into the latent space.

\subsection{Twitter friend recommendation}

For this dataset and the recommendation task, no further hyperparameter tuning is done, values from previous experiments are used.
The parameters of the methods used for comparison are extracted from \cite{Chang18,Chen19}  with their best parameters.
%
The twitter dataset is large with more than 100,000 users which would be intractable for existing methods. Hence, 2506 twitter users are randomly selected from the database as in \cite{Chen19}  for fair comparison.

The 20 most followed accounts (over the whole dataset) are selected, 
For each of these, 10 users following them are chosen (at random) and the average representation from latent space is computed. The latent space is set to dimension $d = 5$. Finally, the cosine similarity is computed between this average profile and the one of the $L = 100$ closest users to these representations. If one of these 100 users actually followed the initially chose account, this is considered as a good friend's recommendation. To assess performance, precision, recall and MRR (mean reciprocal rank) metrics are computed.
The results are in Table \ref{data_results} (right).

The performance of MVGCCA comparable (except for recall) to that of GMCCA, which is currently the best method for this task. 
We recall here that the results are assessed on a limited dataset that all methods can process. However,  the full version with 100,000 users would be intractable for GMCCA, while the proposed MVGCCA scales for this dataset size.

\section{Conclusion}
\label{sec:conclusion}

We proposed MVGCCA, a novel multiview and non linear extension of CCA based on a Bayesian inference model. Our model is scalable  and can take into account  the available graph structural information from the data. The probabilistic nature of the model can be used for addressing missing views or link prediction tasks which we postpone for future work.

\vfill\pagebreak

\bibliographystyle{IEEEbib}
\bibliography{strings,biblio}  

\end{document}